\documentclass{article}

\usepackage{microtype}
\usepackage{graphicx}
\usepackage{subfigure}
\usepackage{booktabs} 

\usepackage{hyperref}

\usepackage{amsmath}

\usepackage[accepted]{icml2018}

\icmltitlerunning{}

\begin{document}

\twocolumn[
\icmltitle{Semantic Segmentation with Scarce Data}


\icmlsetsymbol{equal}{*}

\begin{icmlauthorlist}
\icmlauthor{Isay Katsman}{equal,cornell}
\icmlauthor{Rohun Tripathi}{equal,cornell}
\icmlauthor{Andreas Veit}{cornell}
\icmlauthor{Serge Belongie}{cornell}
\end{icmlauthorlist}

\icmlaffiliation{cornell}{Department of Computer Science, Cornell Tech and Cornell University, NY, USA}
\icmlcorrespondingauthor{Isay Katsman}{isk22@cornell.edu}
\icmlcorrespondingauthor{Rohun Tripathi}{rt443@cornell.edu}

\icmlkeywords{Machine Learning, ICML}

\vskip 0.3in
]



\printAffiliationsAndNotice{\icmlEqualContribution} 

\begin{abstract}
Semantic segmentation is a challenging vision problem that usually necessitates the collection of large amounts of finely annotated data, which is often quite expensive to obtain. Coarsely annotated data provides an interesting alternative as it is usually substantially more cheap. In this work, we present a method to leverage coarsely annotated data along with fine supervision to produce better segmentation results than would be obtained when training using only the fine data. We validate our approach by simulating a scarce data setting with less than 200 low resolution images from the Cityscapes dataset and show that our method substantially outperforms solely training on the fine annotation data by an average of 15.52\% mIoU and outperforms the coarse mask by an average of 5.28\% mIoU.
\end{abstract}

\section{Introduction}

Semantic segmentation is the task of performing per-pixel annotation with a discrete set of semantic class labels. Current state of the art methods in segmentation use deep neural networks \cite{Zhao2017PyramidSP,Shelhamer2015FullyCN} which rely on massive per-pixel labeled datasets \cite{zhou2017scene, Everingham2009ThePV,Cordts2016TheCD}. The labels for these datasets are finely annotated and are generally very expensive to collect. In this paper, we consider the scarce data paradigm, a setting where annotating images with fine details is extremely expensive. On the other hand, coarse or noisy segmentation annotations are substantially cheaper to collect, for example, through use of automated heuristics. 

\begin{figure}
\begin{center}
\includegraphics[width=0.47\textwidth]{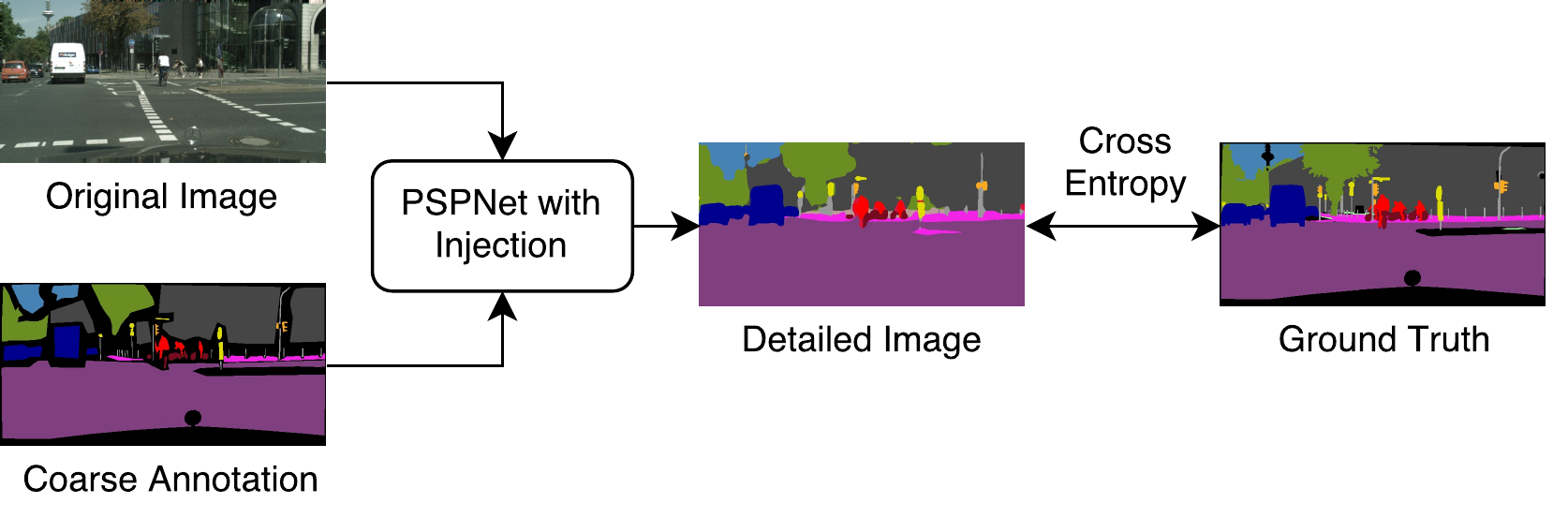}
\caption{\textbf{Overview.} The proposed method is trained to improve the detail of coarse annotations by using them together with regular images in a PSPNet pipeline supervised with fine annotations.}
\label{fig:introduction_figure}
\end{center}
\end{figure}

Recently, Veit et al. \cite{Veit2017LearningFN} introduced a semi-supervised framework that jointly learns from clean and noisy labels to more accurately classify images. They show that the paradigm of first training a network on noisy labels and then fine-tuning on the fine labels is non-optimal. In this work, we propose a related approach for semantic segmentation, in which the goal is to learn jointly from both coarse and fine segmentation masks to provide better image segmentations. 
The proposed model is able to take the input image together with a coarse mask to produce a detailed annotation, Figure~\ref{fig:introduction_figure}. 
In particular, we model the segmentation network as conditionally dependent on the coarse masks, based on the intuition that the coarse masks provide a good starting point for the segmentation task. 

\begin{figure*}
\begin{center}
\includegraphics[width=\textwidth]{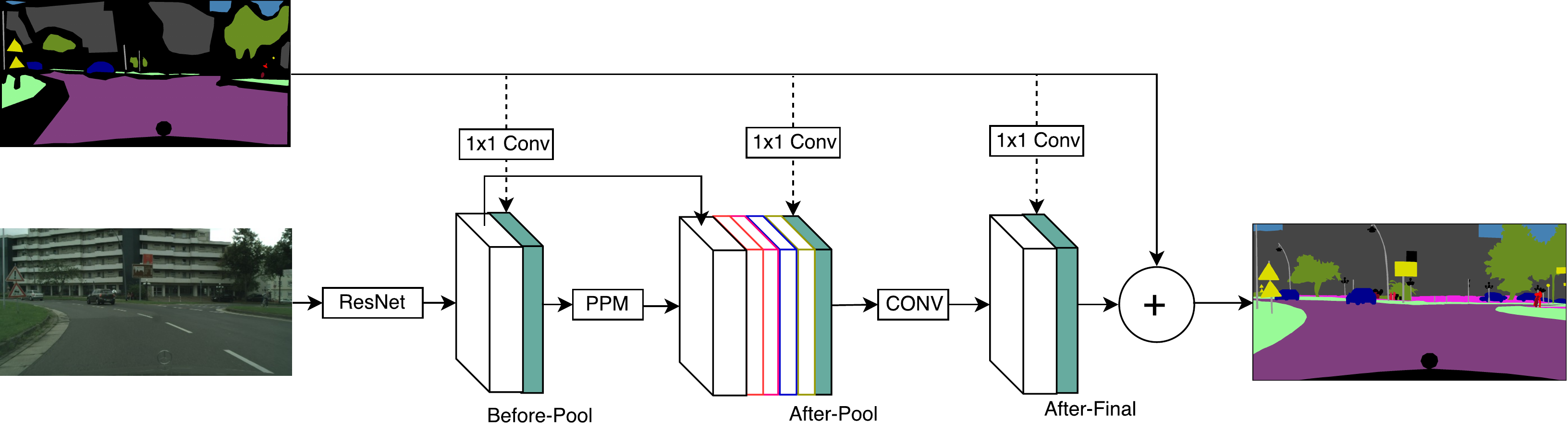}
    \caption{\textbf{Overview of the proposed detailer network.} The solid green boxes represent the possible locations for the injection of the coarse mask embedding. Only one of the injection locations is used for each variant. The architecture is based on the PSPNet.}
\label{fig:insertion_at_different_nodes}
\end{center}
\end{figure*}

We use the Cityscapes dataset to evaluate our method. This segmentation dataset has the important characteristic of providing both fine and coarse annotations. Fine annotations are limited to only 5000 images, whereas coarse annotations are available for these 5000 images and for an additional 20000 images. To simulate a setting of scarce data, we limit our training set to a small number of finely annotated images. We primarily focus on the scarce data setting, which we define as having less than or equal to $200$ finely annotated images (together with their corresponding coarse masks). 

This paper's main contributions are as follows. We present a method for segmentation in the scarce data setting. We utilize the coarse and fine masks jointly and produce better segmentation results than would be obtained when training solely with fine data or using the coarse masks alone. Our approach is validated in a scarce data setting on the Cityscapes dataset, with average gains of 15.52\% mIoU\footnote{Averages are taken over finely annotated dataset sizes of 10, 25, 50, 100, 200.} over the corresponding baseline PSPNet and average gains of 5.28\% over using the coarse mask directly as predictions.

\section{Related Work}
By adapting Convolutional Neural Networks for semantic segmentation, substantial progress has been made in many segmentation benchmarks \cite{Badrinarayanan2017SegNetAD, Ronneberger2015UNetCN, Shelhamer2015FullyCN, Chen2018DeepLabSI, Zhao2017PyramidSP}. Semi-supervised semantic segmentation is an active field of research given that collection of large datasets of finely annotated data is expensive \cite{Pinheiro2016LearningTR, GarciaGarcia2017ARO, Papandreou2015WeaklyandSL}.
In \cite{Pinheiro2016LearningTR}, the authors propose a refinement module to merge rich low level features with high-level object features learned in the upper layers of a CNN. Another method for semi-supervised segmentation is \cite{Papandreou2015WeaklyandSL}, which uses Expectation-Maximization (EM) methods for training DCNN models. 

Providing low level information (such as the coarse masks in our method) in the form of an embedding layer is explored in previous works \cite{Veit2017LearningFN, Perez2017LearningVR}. In \cite{Veit2017LearningFN}, the noisy labels are jointly embedded with the visual features extracted from an Inception-V3 \cite{Szegedy2016RethinkingTI} ConvNet. We explore an analogous approach, which concatenates the coarse masks with the convolution blocks at different network locations in the scarce data semantic segmentation domain.

\begin{figure*}
\begin{center}
\includegraphics[width=\textwidth]{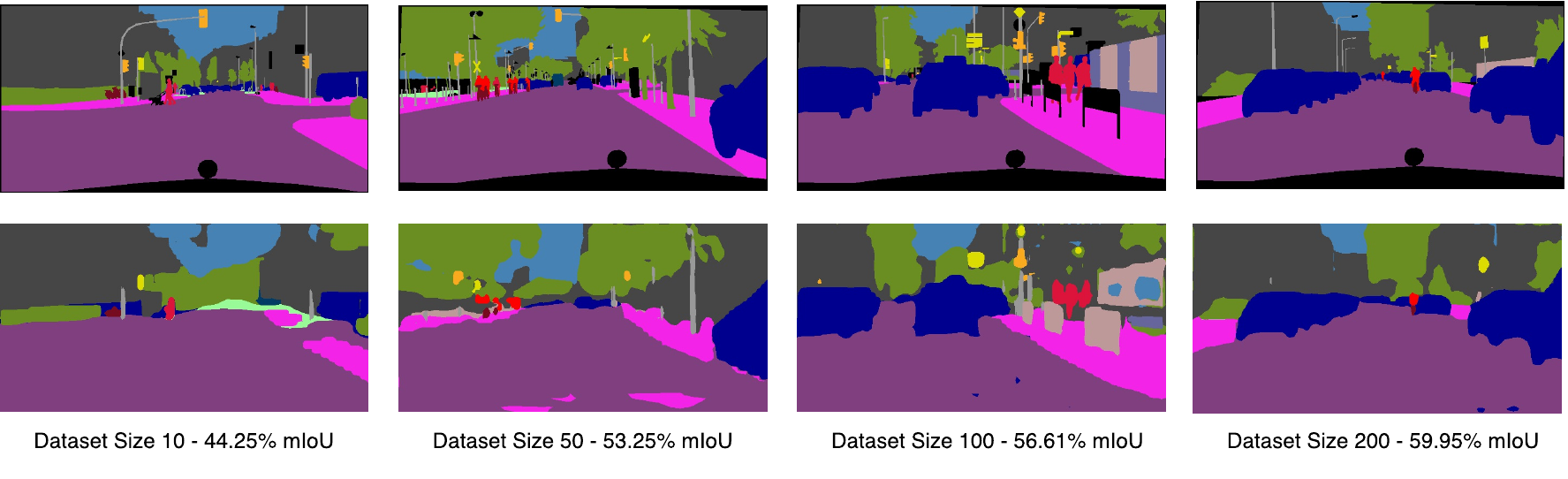}
    \caption{\textbf{Qualitative results of the proposed method} on Cityscapes ~\cite{Cordts2016TheCD}. Top images show the ground truth and the bottom images show the result of our model. The models used for predicted images are trained on 10, 50, 100, and 200 images with a low resolution of $340\times 340$. Pixels that correspond to black pixels in the ground truth are ignored during validation, as 'ignore labels'.}
\label{fig:qualitative_results}
\end{center}
\end{figure*}

\section{Approach}
Our goal is to leverage the small set of fine annotations and their coarse counterparts to learn a mapping from the coarse to fine annotation space. Both the coarse and fine segmentation labels consist of 20 class labels (one being the void label). Formally, we have a small dataset $V$ of triplets, each of which consists of fine annotation $v$, coarse annotation $y$ and image $m$. We thus have
$V = \{(v_{i}, y_{i}, m_{i}), ...\}$. Each annotation is a mask, with pixels that can take 20 values (each value representing a different class). 
Our goal is to design an effective approach to leverage the quality of the labels in $v$ to map from the coarse labels in $y$ to improved segmentation masks. The mapping is conditioned on the input image. We coin the name ``detailer'' for an architecture that achieves this goal.

\subsection{Architecture of Detailer}
\label{architecture}

We utilize a PSPNet for our classifier baseline with training details as described in the original paper. We propose a detailer architecture that is a modified PSPNet, depicted in Figure \ref{fig:insertion_at_different_nodes}. For each triplet in $V$, we forward $m_{i}$ through the network and insert an embedding of the coarse annotation, $y_{i}$, into the network. The embedding is inserted either before the PPM module (before-pool), after the PPM module (after-pool) or after the final convolution block (after-final). In Section \ref{results}, Results, we use the after-final embedding\footnote{This decision is justified with an ablation study. See Appendix.}.

To embed the coarse mask, we expand the 2D annotation image of size $W \times H$ to a $19\times W \times H$ tensor, where $W$ and $H$ are the height and width of the mask and each channel is a binary $W\times H$ mask. The channel-wise binary vector for each pixel is a one-hot encoding for each pixel, i.e., it takes 1 at the $k^{th}$ index, where $k$ is the class of the pixel. A 1x1 convolution is applied on this mask tensor to produce its embedding. The number of filters used for this function is a hyper-parameter, with the default value being 800 filters\footnote{This decision is made via an ablation study. See Appendix.}. This embedding is concatenated with the visual feature tensor at the configured network location. 

Another key detail of the detailer network is an identity-skip connection that adds the coarse annotation labels from the training set to the output of the cleaning module. The skip connection is inspired by the approach from \cite{He2016DeepRL}. Due to this connection, the network only needs to learn the difference between the fine and the coarse annotations labels, thus simplifying the learning task. For the $i^{th}$ image, the coarse mask, $y_{i}$ and the predicted image annotation tensor, $p_{i}$, both take the shape $19\times W \times H$. The first denotes a one-hot encoding tensor described earlier and the second represents the corrections for the predicted class values. These two tensors are added together for the final detailed prediction.

\section{Experimental Setting}
Our experiments are done with simulated scarce subsets of the Cityscapes dataset, using mIoU as our evaluation metric.

\subsection{Dataset}
We evaluate our proposed architecture on the Urban Landscape Cityscapes Dataset ~\cite{Cordts2016TheCD}. The dataset is suited for our task as it contains a coarse annotation along with a fine annotation for 5000 images. The coarse mask is made with an objective to annotate as many pixels as possible within 7 minutes. This was achieved by labeling coarse polygons under the sole constraint that each polygon must only include pixels belonging to a single class. These coarse masks have very low false positives, 97\% of all labeled pixels in the coarse annotations were assigned the same class as the fine annotations.
In comparison, the annotation process of each of the fine masks takes more than 1.5 hours on average. 

The mIoU of the coarse masks compared against the fine masks for the validation set is $47.81\%$. Because our setting uses the coarse masks at test time, an mIoU of $47.81\%$ can be achieved by directly outputting the coarse mask. Thus, our detailer result should exceed this value at test time.

\subsection{Training Details}
\label{sec:training_details}
We use data augmentation as follows - random sizing (between 0.5 and 2.0), random cropping, random rotation (between -10 and +10 degrees) and random horizontal flip. Resnet-101 pretrained on ImageNet was used to initialize the initial layers of the PSPNet. Pixel values are normalized using the ImageNet mean and standard deviation.
To train the networks we use cross-entropy loss and SGD with an initial learning rate 0.01, polynomial learning rate decay of 0.9 every iteration, and a momentum of 0.99. We stopped training after 10K mini-batches (batch size 8) for datasets sizes of 200 and under, and after 90k mini-batches for dataset sizes above 200 (to allow training to fully converge).

\section{Results} \label{results}

Our approach shows that for dataset sizes that are very small, it is still possible to learn a segmentation network with relatively high performance. We show that our detailer outperforms both the course mask and the regular classifier for small dataset sizes. Surprisingly, we also notice that although higher resolution images typically work much better than low resolution images when a lot of segmentation data is available, the same does not hold true in the scarce data setting for the detailer (elaboration in ``Higher Resolution'' section).

When utilizing a very small dataset of fine and coarse annotations to train, our detailer network performs reasonably well on the Cityscapes validation set, as shown in Table \ref{table:diff_size_340}. For every size, the detailer outperforms the classifier. The average gain is 15.52\% mIoU over the classifier performance. Starting from only a dataset of 10 images up to 200 images, we also outperform the coarse mask by an average of 5.28\% (the coarse mask mIoU is 47.81\%). Qualitative results are shown in Figure \ref{fig:qualitative_results}.

\begin{table}[t]
\vskip 0.15in
\begin{center}
\begin{tabular}{ccc}
\toprule
Dataset size & Detailer mIoU & Classifier mIoU\\
\midrule
10    & 44.25\% & 23.54\% \\
25    & 51.34\% & 35.49\% \\
50    & 53.26\% & 38.54\% \\
100 & 56.61\% & 42.75\% \\
200 & 59.95\% & 47.50\%\\
500 & 60.45\% & 45.78\%\\
2975 (complete) & 64.94\% & 64.11\%\\
\bottomrule
\end{tabular}
\end{center}
\caption{\textbf{Segmentation performance} for the proposed detailer and PSPNet with varying dataset size. Models are trained with an image resolution of $340 \times 340$. The proposed detailer clearly outperforms the standard PSPNet for all dataset sizes with the largest improvements in the regime with very few training examples.}
\label{table:diff_size_340}
\vskip -0.1in
\end{table}

\textbf{Composite Prediction}. A reasonable concern is that our improvement over the baseline classifier model shown in Table \ref{table:diff_size_340} is solely due to the fact that we incorporate the coarse mask (which is already at 47.81\%). To confirm that our coarse injection model (Figure \ref{fig:insertion_at_different_nodes}) improves results, we define the notion of a composite model. A composite model takes the coarse mask and adds the predictions of the trained segmentation model to it for all pixels that were not assigned a label by the coarse mask (pixels which had 'ignore'). From Table \ref{tab:higher-res} we see that if we take the trained detailer and classifier and make them composite, the detailer outperforms the classifier, indicating that the injection model approach does in fact  improve performance.

\begin{table}
\begin{center}
\begin{tabular}{ ccccc }
\toprule
Model & mIoU \\ 
\midrule
Coarse Mask & 47.81\% \\ 
Classifier & 42.75\% \\
Detailer & 56.61\% \\
Classifier Composite & 58.08\% \\
Detailer Composite & 60.60\% \\
\bottomrule
\end{tabular}
\end{center}\caption{\textbf{mIoU for different models}. We compare models trained on images with resolution $340 \times 340$ with a dataset size of 100. We observe that the detailer outperforms the classifier in the standard and the composite settings. }
\label{tab:higher-res}
\end{table}

\textbf{Higher Resolution}. As opposed to using a scarce data low-resolution setting, it is also presumable that a slightly more expensive setting of scarce data with high-resolution could occur in the real-world. To simulate this, we use the same approach as before but instead of re-sizing the Cityscapes data to $340 \times 340$, we re-size to $850 \times 850$. If we compare the detailer results at both resolutions (Table \ref{tab:higher-res}), we see that for most dataset sizes, at low resolution the detailer does better than at high resolution. 
This is an important result as it shows that in the scarce data setting while using a coarse mask, using low resolution annotations is preferable to high resolution annotations. Our results indicate that one can save resources by providing coarse and fine annotations for a low resolution $340 \times 340$ instead of providing annotations at a high resolution of $850 \times 850$.

\begin{table}
\begin{center}
\begin{tabular}{ ccccc }
\toprule
Dataset Size & Detailer (higher res) & Detailer (lower res) \\ 
\midrule
10 & 40.91\% & 44.20\% \\ 
25 & 46.19\% & 52.33\% \\
50 & 48.67\% & 51.51\% \\
100 & 52.87\% & 53.96\% \\
200 & 57.88\% & 56.27\% \\
500 & 59.83\% & 58.10\% \\
\bottomrule
\end{tabular}
\end{center}\caption{\textbf{mIoU for varying dataset size for detailer networks}. We compare models trained on images with resolutions $340 \times 340$ and $850 \times 850$ while varying the number of training examples from 10 to 500. We observe that the former outperforms the latter when trained with lower number of images and the latter performs better as the number of data points exceeds 200.}
\label{tab:higher-res}
\end{table}

\textbf{Distillation} Our detailer predicts detailed masks using coarse masks as input. These detailed masks can be used to distill the semantic information to a student network, thereby removing our need for coarse masks at test time. We observe that training a PSPNet network using just 100 detailed masks\footnote{These detailed masks are obtained from our detailer trained on a scarce dataset of 100 images.} gives 37.85\% mIoU, exceeding the 34.61\% mIoU obtained when training a PSPNet with the corresponding 100 coarse masks. This result shows that the detailing method captures important semantic information which can be transfered to a student network.

\section{Conclusion}
We propose a method of leveraging coarse annotations to improve performance of deep neural networks for semantic segmentation in the scarce data setting. Our results show that in the case of scarce data, significant improvement can be attained with the addition of coarse annotations. We observe an increase in segmentation performance of 15.52\% mIoU on average when comparing the proposed detailer to a vanilla PSPNet. Additionally, we observe that although models trained on higher resolution images tend to perform better given sufficient training data, in the setting of scarce training data, low resolution images lead to better performance.
\bibliography{example_paper}
\bibliographystyle{icml2018}

\clearpage
\section{Appendix}

We perform ablation studies to justify our architectural decisions for the detailer. The After-Final location is selected for coarse embedding injection based on the results of the ablation study shown in Table \ref{tab:insertion-node}. We choose 800 filters for the coarse embedding in the detailer network based on the ablation results in Table \ref{tab:filter-count}.

\begin{table}[!ht]
\begin{center}
\begin{tabular}{ ccccc }
\toprule
Coarse Location Insertion & mIoU \\ 
\midrule
\textbf{After Final} & \textbf{52.81\%} \\ 
Before Pool & 41.06\% \\
After Pool & 44.09\%  \\
\bottomrule
\end{tabular}
\end{center}\caption{\textbf{mIoU for various coarse information embedding locations.} The detailer network is trained on 100 images of resolution $340 \times 340$. Note that the corresponding classifier result for this resolution and dataset size is 42.75\% mIoU.} 
\label{tab:insertion-node}
\end{table}

\begin{table}[!ht]
\begin{center}
\begin{tabular}{ ccccc }
\toprule
Embedding Size & mIoU \\ 
\midrule
19 & 56.26\%  \\ 
600 & 58.98\% \\
700 & 58.51\% \\
750 & 59.19\% \\
\textbf{800} & \textbf{59.98\%} \\
850 & 59.21\%\\
1000 & 59.95\% \\
1500 & 58.78\% \\
\bottomrule 
\end{tabular}
\end{center}\caption{\textbf{mIoU for various sizes of coarse embedding.} The detailer network is trained on 200 images of resolution $340 \times 340$. We observe that performance increases until an embedding size of 800. For larger embedding sizes, the model starts over-fitting. Note that the performance is significantly above the 47.51\% mIoU of the classifier for all embedding sizes.}
\label{tab:filter-count}
\end{table}
\end{document}